# IMPROVING AGE PREDICTION: UTILIZING LSTM-BASED DYNAMIC FORECASTING FOR DATA AUGMENTATION IN MULTIVARIATE TIME SERIES ANALYSIS

*Yutong Gao[1], Charles A. Ellis[1], Vince D. Calhoun[1] and Robyn L. Miller[1]*
[1]*Tri-Institutional Center for Translational Research in Neuroimaging and Data Science (TReNDS):*
Georgia State University, Georgia Institute of Technology, Emory University, Atlanta, Georgia

## ABSTRACT

The high dimensionality and complexity of neuroimaging data necessitate large datasets to develop robust and high-performing deep learning models. However, the neuroimaging field is notably hampered by the scarcity of such datasets. In this work, we proposed a data augmentation and validation framework that utilizes dynamic forecasting with Long Short-Term Memory (LSTM) networks to enrich datasets. We extended multivariate time series data by predicting the time courses of independent component networks (ICNs) in both one-step and recursive configurations. The effectiveness of these augmented datasets was then compared with the original data using various deep learning models designed for chronological age prediction tasks. The results suggest that our approach improves model performance, providing a robust solution to overcome the challenges presented by the limited size of neuroimaging datasets.

## 1. INTRODUCTION

Deep learning has garnered significant attention in the domain of neuroimaging for classification and prediction [1]. Given the temporal and spatial complexity of brain activity, training deep learning models typically requires large datasets. However, for many neuroimaging tasks, especially those related to mental disorders, only relatively small datasets are available. For example, data for Alzheimer's Disease, the fifth leading cause of death [2], has been collected in major public datasets like OASIS3 [3] and ADNI (http://adni.loni.usc.edu), but those datasets have only a few hundred subjects in disease states. Such small or imbalances in dataset size and sequence length can lead to model overfitting [4], resulting in poor generalization. The creation of large, public datasets for various mental disorders and their various stages remains a challenge, as does subjecting individuals to extended scanning sessions to acquire longer scans.

Data augmentation presents a viable solution for enhancing the diversity of training datasets without the need for additional data collection, with the goal of achieving improved task performance [5]. This process involves applying a series of transformations, such as rotation, cropping, and noise injection for imaging data, as well as jittering, scaling, and time-warping for time-series data. [4] details the application of resampling techniques to a small fMRI dataset. While such transformations are cost-effective, they may be constrained by the quality of the training set and may not preserve the temporal dynamics inherent in time-series data. Data augmentation can also be facilitated through deep learning models, such as Generative Adversarial Networks (GANs), which are capable of generating synthetic fMRI data [6], Additionally, training Recurrent Neural Networks (RNNs) to dynamically predict future states serves as another method for data augmentation, as demonstrated by this work. Model-based augmentation through deep learning can potentially learn and replicate underlying data patterns, with RNN-based models being particularly effective at capturing and modeling temporal dependencies for improved prediction of future states.

In this study, we applied dynamic forecasting to enhance the multivariate time series data by employing both one-step and recursive LSTM networks to predict the time courses of ICNs. We first assessed the efficacy of these augmented time courses during the augmentation phase. Subsequently, we performed an empirical comparison of the augmented dataset, which incorporated dynamic forecasting, with the original dataset. This comparison utilized various deep learning models, such as multi-channel CNN, multi-channel CNNs with attention mechanisms, and Time-Attention LSTMs, specifically for a chronological age prediction task. Our findings indicate that the augmentation approach improves the performance of the prediction task when compared to using the original dataset alone. The overview of the work pipeline is shown in Figure 1.

## 2. MATERIRALS AND METHODS

### 2.1. Dataset

We evaluated our method on UK Biobank Brain Imaging [7], which comprises high-quality imaging data from healthy subjects. We utilized a dataset encompassing 7,025 rs-fMRI scans, with the participants' ages averaging at 59.17 with a standard deviation of 4.87 years. Similar to [8], we preprocessed the rs-fMRI data by employing group independent component analysis (GICA) using the NeuroMark pipeline [9] to extract 53 ICNs across seven

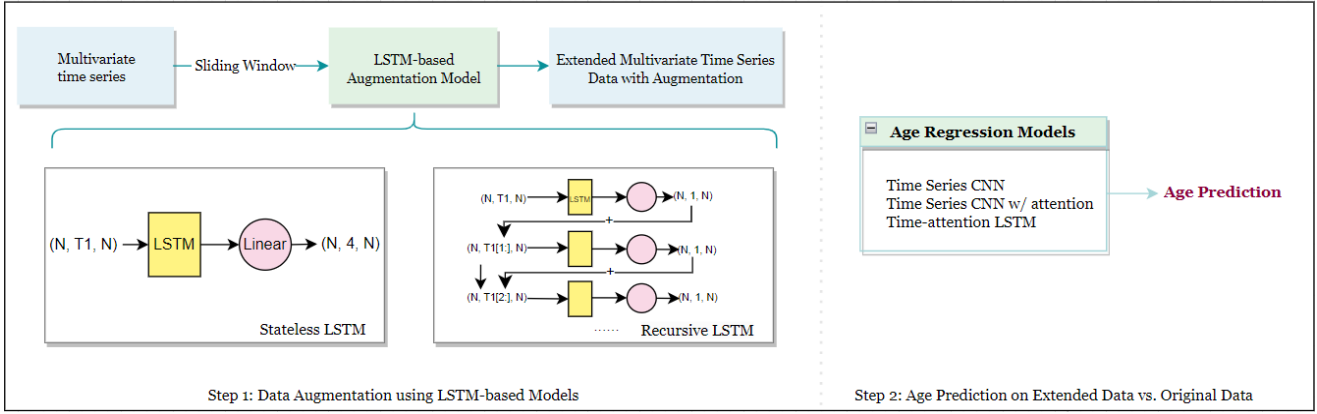

Figure 1 provides an overview of the work pipeline. The first step in this work is forecasting future time periods as a means of temporal dimension augmentation. Two types of models were involved in this step: a stateless LSTM, which forecasts four time points in one iteration, and a recursive LSTM, which iteratively forecasts one time point at each iteration, with a recursion depth of four. The second step is the validation of the augmentation stage, which involves three models: a Time Series CNN, a Time Series CNN with Attention, and a Time-Attention LSTM. These models are used to independently train on the original and augmented datasets for an age prediction task.

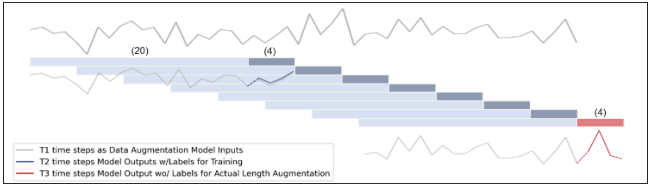

Figure 2 The sliding window technique segments the time series data. The full-length time courses for each ICN are divided into segments of length 24. Of these, T1, which consists of 20 time points, is used for training the forecasting model, and T2, which consists of 4 time points, represents the forecasted results that are compared with the ground truth. T3, indicated in red, is the final augmented portion, the length of which depends on the chosen augmentation model.

functional domains including subcortical, visual, auditory, sensorimotor, cognitive control, default mode, and cerebellar networks. Subsequently, the derived time courses of these ICNs were subject to downsampling. We reduced the temporal resolution from a 0.735-time repetition (TR) to a 2.94 TR, which resulted in a total of 122 time points for each of the 53 channels. We adjusted the temporal resolution to align the 0.735-second TR with those of popular datasets that have been used in other resting-state fMRI studies [3] and to which our augmentation approach might conceivably be applied in the future. We sought to ensure that our pipeline would be generalizable to other datasets and tasks, particularly those that most affected by the limitations of small datasets.

### 2.2. Sliding Window Prepared for Data Augmentation

The segmentation of preprocessed ICN time courses for data augmentation was executed using a sliding window approach. Each window encompassed a segment length of 24 time points, with the initial 20 time points serving as the training data for forecasting, and the subsequent 4 time points acting as the label for the forecasting model. Figure 2 illustrates the process. The selection of window size allows for maximization of data utilization for the available time courses. The final segment window is used for forecasting the extended portion of the dataset for purposes of data augmentation, which highlighted in red within the figure.

### 2.3. Data Augmentation Model

*Stateless LSTM*- The model consists of a single LSTM layer with a hidden size of 50, followed by a linear output layer. A key implementation detail is the resetting of hidden states to zero after processing each batch. This strategy is adopted because our data segments overlap, and it is crucial to ensure that the model does not retain memory of sequences from distinct batches, thereby preventing interference between them. This same configuration is also applied in the training of the recursive LSTM model

*Recursive LSTM*- The model consists a single LSTM layer with a hidden size of 50, coupled with a linear output layer. The distinct feature of our approach lies in the forecasting method: for each forecasting instance, only a single time point for all the ICNs is predicted. This prediction, along with the previous time points excluding the first ($T[1:]$), is then utilized to forecast the subsequent time point. This process is repeated until all four steps of forecasting are completed during training. Note that the step size for predictions is adaptable to the needs of the model's forecasting phase.

### 2.4. Age Prediction Model

Three deep learning models were assessed using baseline and forecasting-augmented datasets for extending multivariate time series. We utilized trained stateless LSTM and recursive LSTM models with various step sizes. The performance was evaluated with the Mean Absolute Error (MAE) serving as the metric for evaluation, which clearly demonstrates the model's capability in predicting age with respect to error performance.

*Time Series CNN*: The model comprises 53 channels. For each channel, the first layer in each channel is a 1D convolutional layer with 128 filters, followed by a second 1D convolutional layer with 64 filters, both having a kernel size of 3. After each convolutional layer, a rectified linear unit (ReLU) serves as the activation function, succeeded by a max pooling layer by a max pooling layer with a kernel size of 2, stride of 2, and padding of 1. Concluding this sequence, two fully connected layers are concatenated to form the final output of the model.

*Time Series CNN with attention-* the model uses the time series CNN as its backbone. A multi-head attention layer [10], with the head parameter set at 2, is strategically positioned after the second max pooling layer and before the first fully connected layer.

*Time-Attention LSTM-* The model was introduced in [8] and is very efficient at classifying brain imaging-related multivariate time series. The model consists of three LSTM layers, each with a hidden size of 64. The time attention layer employs scaled dot-product attention [10] with the attention-weighted output $c_t$ calculated as shown in equation (1). The attention output, $c_t$, is subsequently reduced to one element at a time by the time attention layer [8].

$$c_t = \frac{1}{\|e_j\|}\sum_{j=1}^{j} \alpha_{tt} \times v_{tj} \quad and \quad \alpha_{tt} = \sigma\left(\frac{q_{tj} \times e_{tj}^\top}{\sqrt{\|e_j\|}}\right) \quad (1)$$

### 2.5. Experiment Settings

During the data augmentation stage, the preprocessed data were segmented into windows. The segmented data were then divided into training and test datasets at the subject level, using an 80:20 ratio. Two augmentation modes, Stateless LSTM and Recursive LSTM, were trained and tested, and the model parameters were saved for further validation. During the age prediction stage, we performed dynamic forecasting using the trained Stateless LSTM to construct the extended multivariate time series ICN data. We evaluated both the original and extended datasets using the Time-series CNN, Time-series CNN with attention, and Time Attention LSTM to first assess whether the performance improved with the extended dataset and to identify the optimal age prediction model. Following the selection of the optimal model from the previous step, we utilized the trained Recursive LSTM to assess the effects of differing extended step sizes, ranging from 4 to 14 in increments of 2. To ensure robust evaluation, a ten-fold cross-validation approach was implemented, and the resulting mean and standard deviation of the MAE are reported as metrics.

## 3. RESULTS AND DISCUSSIONS

### 3.1. Training the Dynamic Forecasting Models

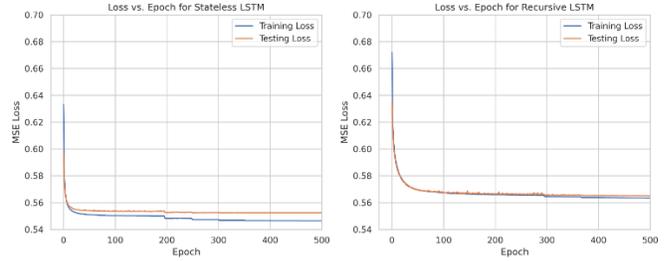

Figure 3 Loss versus Epoch for Data Augmentation Training: The left plot corresponds to the stateless LSTM model, and the right plot to the recursive LSTM model.

Figure 3 displays the training and testing loss curve over 500 epochs, providing insights into the models learning progress and stability. The left plot shows mean squared error (MSE) with each epoch for forecasting four-time steps at once using the Stateless LSTM, whereas the right plot shows the MSE across epochs for recursive four-step predictions with Recursive LSTM. The stable and consistent performance across training and testing datasets implies that the models have not overlearned from the training data and can be expected to make reliable predictions on new data.

### 3.2. Identification of Optimal Age Prediction Model

Table 1 provides a comparative analysis of the Mean Absolute Error (MAE) across various age prediction models, including the Time Series CNN, Time Series CNN with Attention, and the Time-Attention LSTM, evaluating their performance on both the baseline and augmented datasets. The findings reported in Table 1 indicate that the models trained on the augmented dataset achieved superior performance compared to those trained on the baseline dataset. Among the three deep learning architectures evaluated, the Time-Attention LSTM demonstrated the most proficient performance. It outperformed the other two CNN-based models in terms of predictive accuracy. This suggests that the model is particularly adept at handling temporal dependencies and learning from multivariate time series data for age prediction.

### 3.3. Evaluation of Forecasting with Varying Step Size

Subsequent to our primary analysis, we assessed the performance of the Recursive LSTM across a range of forecasting steps, the outcomes of which are summarized in Table 2. The augmented data from this evaluation strongly suggests that models trained on the augmented dataset consistently surpassed the performance of those trained on the baseline dataset, irrespective of the step size utilized. Notably, it was observed that a step size of 10 yielded the optimal MAE, marking it as the most effective forecasting interval among the range tested.

### 3.4. Limitation and Future Works

The preliminary outcomes of this study suggest that generative forecasting models could be a viable strategy for augmenting deep learning tasks, particularly when there is a scarcity of training data. Our approach has the potential to expand datasets by generating extended time courses and to amplify the data size by segmenting the generated data into shorter lengths. Although our proposed work offers a solution of data augmentation with dynamic forecasting, it also opens up new areas for investigation, such as determining the optimal segmentation length for training sequences. Future work in generative modeling could broaden its applications, filling in missing values or channels and unraveling the dynamic principles that underpin data behavior, which could significantly aid interpretability efforts.

Table 1 MAE Comparison of Age Prediction Models (Timeseries CNN, Timeseries CNN w/att, Time-Attention LSTM) Performance before and after augmentation with stateless LSTM.

|  | Timeseries CNN | Timeseries CNN w/ Att | Time-Att LSTM |
| --- | --- | --- | --- |
| Pre-Aug. | 4.1376 | 4.1475 | 4.0303 |
| After-Aug. | 4.1305 | 4.1334 | **4.0228** |

Table 2 MAE Comparison of Age Prediction with model Time-Att LSTM before and after augmentation with recursive LSTM with different steps

| Time-Att LSTM | Pre-Aug. | After-Aug. |
| --- | --- | --- |
| Baseline | 4.0303±0.1195 | - |
| Step 4 | - | 4.0229±0.1164 |
| Step 6 | - | 4.0214±0.1183 |
| Step 8 | - | 4.0216±0.1158 |
| Step 10 | - | **4.0199**±0.1159 |
| Step 12 | - | 4.0240±0.1167 |
| Step 14 | - | 4.0227±0.1165 |

## 4. CONCLUSIONS

In this study, we address the challenge of insufficient neuroimaging datasets for deep learning training. We implemented two types of LSTM networks to predict the future states of ICNs derived from rs-fMRI. The stateless LSTM offers a cost-efficient forecasting approach by predicting n time steps at single iteration. The second model, the recursive LSTM, provides greater flexibility to forecast one step at once and recursively use the forecasted steps to process the subsequent iteration. Our finding indicates that forecasting over longer time courses can enhance the accuracy of age estimation tasks. Beyond the immediate application, these findings could potentially be adapted for a broader range of applications where data availability is a significant constraint.

## 5. ACKNOWLEDGMENT

Research supported by NSF 2112455.## 6. REFERENCES

[1] W. Yan *et al.*, "Deep learning in neuroimaging: Promises and challenges," *IEEE Signal Processing Magazine,* vol. 39, no. 2, pp. 87-98, 2022.

[2] A. s. Association, "2018 Alzheimer's disease facts and figures," *Alzheimer's & Dementia,* vol. 14, no. 3, pp. 367-429, 2018.

[3] P. J. LaMontagne *et al.*, "OASIS-3: Longitudinal Neuroimaging, Clinical, and Cognitive Dataset for Normal Aging and Alzheimer Disease," *medRxiv,* p. 2019.12.13.19014902, 2019, doi: 10.1101/2019.12.13.19014902.

[4] N. C. Dvornek, D. Yang, P. Ventola, and J. S. Duncan, "Learning generalizable recurrent neural networks from small task-fmri datasets," in *International Conference on Medical Image Computing and Computer-Assisted Intervention*, 2018: Springer, pp. 329-337.

[5] C. A. Ellis, R. L. Miller, and V. D. Calhoun, "Pairing Explainable Deep Learning Classification with Clustering to Uncover Effects of Schizophrenia Upon Whole Brain Functional Network Connectivity Dynamics," *bioRxiv,* p. 2023.03. 01.530708, 2023.

[6] N. Qiang *et al.*, "Functional brain network identification and fMRI augmentation using a VAE-GAN framework," *Computers in Biology and Medicine,* vol. 165, p. 107395, 2023.

[7] K. L. Miller *et al.*, "Multimodal population brain imaging in the UK Biobank prospective epidemiological study," *Nature neuroscience,* vol. 19, no. 11, pp. 1523-1536, 2016.

[8] Y. Gao, N. Lewis, V. D. Calhoun, and R. L. Miller, "Interpretable LSTM model reveals transiently-realized patterns of dynamic brain connectivity that predict patient deterioration or recovery from very mild cognitive impairment," *Computers in Biology and Medicine,* vol. 161, p. 107005, 2023.

[9] Y. Du *et al.*, "NeuroMark: An automated and adaptive ICA based pipeline to identify reproducible fMRI markers of brain disorders," *Neuroimage Clin,* vol. 28, p. 102375, 2020, doi: 10.1016/j.nicl.2020.102375.

[10] A. Vaswani *et al.*, "Attention is all you need," *Advances in neural information processing systems,* vol. 30, 2017.© 2023 IEEE Preprint Under Review